\documentclass[10pt, a4paper]{article}

\usepackage[final]{lrec-coling2024}
\usepackage{booktabs}
\usepackage{enumitem}
\usepackage{tcolorbox}
\usepackage{subcaption}

\title{TartuNLP at EvaLatin 2024: Emotion Polarity Detection}

\name{Aleksei Dorkin, Kairit Sirts} 

\address{University of Tartu \\
         Tartu, Estonia \\
         \{aleksei.dorkin, kairit.sirts\}@ut.ee\\}

\abstract{
This paper presents the TartuNLP team submission to EvaLatin 2024 shared task of the emotion polarity detection for historical Latin texts. Our system relies on two distinct approaches to annotating training data for supervised learning: 1) creating heuristics-based labels by adopting the polarity lexicon provided by the organizers and 2) generating labels with GPT4. We employed parameter efficient fine-tuning using the adapters framework and experimented with both monolingual and cross-lingual knowledge transfer for training language and task adapters. Our submission with the LLM-generated labels achieved the overall first place in the emotion polarity detection task. Our results show that LLM-based annotations show promising results on texts in Latin.
 \\ \newline  }

\begin{document}

\maketitleabstract

\section{Introduction}

This short report describes the system developed the TartuNLP team for the Emotion Polarity Detection task of the EvaLatin 2024 Evaluation Campaign~\cite{sprugnoli-etal-2024-overview}. 
The goal of the task was to label Latin texts from three historical authors with four emotion polarity labels as positive, negative, neutral or mixed. For this task, no training data was provided, but only a polarity lexicon and a small evaluation set with 44 annotated sentences.

Our approach entails two steps. First, we annotated data for supervised model training a) via heuristic rules using the provided polarity lexicon and b) using GPT-4 (see Section \ref{sec:data}). Secondly, we adopted knowledge transfer with parameter-efficient training via adapters \citep{houlsby2019parameter} followed by task-specific fine-tuning on the data annotated in the first step (see Section \ref{sec:system}). The knowledge transfer was applied both cross-lingually via pretraining on an English sentiment analysis task, and monolingually by training on an unannotated Latin text corpus. 

We made two submissions to the shared task: one with heuristically annotated training data and another with the GPT-4 annotated labels. Both submissions obtained competitive results, with the submission with GPT-4 labels obtaining the first place overall. The code for the system is available on GitHub.\footnote{\url{https://github.com/slowwavesleep/ancient-lang-adapters/tree/lt4hala}}

\section{Data Annotation}
\label{sec:data}

For the Emotion Polarity Detection task, no training data was provided. However, the organizers provided two useful resources: a polarity lexicon and a small gold annotated sample. We employed two distinct approaches to annotate the training data based on these resources: a heuristics-based and an LLM-based. The annotated data from both approaches is available on HuggingFace Hub.\footnote{\url{https://huggingface.co/datasets/adorkin/evalatin2024}} The label distribution for the annotated data is presented in Table \ref{tab:stats}.

\subsection{Heuristics-based annotation}

\begin{table}[t]
\centering
\begin{tabular}{lcc}
\toprule
\textbf{Label} & \textbf{Heuristics} & \textbf{LLM-based} \\
\midrule
positive & 6535 & 1334 \\
negative & 2243 & 1028 \\
mixed     & 5884 & 221 \\
neutral     & 735 & 4698 \\
\midrule
Total & 15396 & 7281 \\
\bottomrule
\end{tabular}
\caption{Statistics of the annotated training data.}
\label{tab:stats}
\end{table}

In this approach, we employed the provided polarity lexicon similarly to the lexicon-based classifier by \citet{sprugnoli2023sentiment}. First, data from all available Universal Dependencies~\cite{11234/1-5150} sources (Version 2.13, the most recent one at the time of writing) in Latin was collected :

\begin{enumerate}[label=\arabic*),noitemsep]
    \item Index Thomisticus Treebank (ITTB);
    \item Late Latin Charter Treebank (LLCT);
    \item UDante;
    \item Perseus;
    \item PROIEL treebank.
\end{enumerate}

Then, the sentences containing no nouns or adjectives in the lexicon were removed. The filtered sentences were assigned labels based on the following rules:

\begin{enumerate}[label=\arabic*),noitemsep]
    \item If all words in the sentence are neutral according to the polarity lexicon, the sentence was labeled as neutral;
    \item If the mean polarity of the words in the sentence is in the range from -0.1 to 0.1, then the sentence was labeled as mixed;
    \item If the mean polarity is larger than 0.1, then the sentence was labeled as positive;
    \item If the mean polarity is less than 0.1, then the sentence was labeled as negative.
\end{enumerate}

Our expectation from this approach was that training a model on lexicon-annotated data would result in a model with better generalization capabilities than simply applying the lexicon classifier.
The total amount of sentences annotated this way was 15396.

\begin{table*}[t]
\centering
\begin{tabular}{lcc}
\toprule
\textbf{Model} & \textbf{Micro Average F1} & \textbf{Macro Average F1} \\
\midrule
TartuNLP\_2     & \textbf{0.34} & \textbf{0.29} \\
TartuNLP\_1     & 0.32 & 0.27 \\
NostraDomina\_1 & 0.22 & 0.28 \\
NostraDomina\_2 & 0.22 & 0.22 \\
\bottomrule
\end{tabular}
\caption{The overall results of all teams.}
\label{tab:results}
\end{table*}

\subsection{LLM-based annotation}
In this approach, we made use of the OpenAI's GPT-4 model via the API (\texttt{gpt-4-turbo-preview}\footnote{\url{https://platform.openai.com/docs/models/gpt-4-and-gpt-4-turbo}}). The sentences were again sampled from the Universal Dependencies sources. The model was given the description of the problem and one example per label from the gold annotations file. The model was tasked with assigning the given sentence a label and providing an explanation as to why it assigned that particular label.

With this approach, we expected that GPT-4 could simulate the annotation process done by an expert in Latin. According to the first author's somewhat limited understanding of Latin and based on a small sample of annotations and explanations done by the model, the output seems reasonable.
We set out to spend about 15 euros per data annotation, which after removing sentences with invalid labels resulted in 7281 annotated sentences.

\section{Description of the system}
\label{sec:system}

The system in our submission is based on the BERT architecture~\cite{devlin-etal-2019-bert}. More specifically, we employed the multilingual version of RoBERTa~\cite{liu2019roberta}---XLM-RoBERTa~\cite{conneau2020unsupervised}, which was trained on the data that included Latin. 

We treated Emotion Polarity Detection as a multi-class classification problem and fine-tuned the model accordingly. However, instead of full fine-tuning, we trained a stack of adapters: a language adapter and a task adapter. Training adapters involves adding a small number of trainable parameters to the model while freezing the rest of the parameters~\cite{houlsby2019parameter}. In addition to making the training considerably faster, adapters mitigate overfitting and catastrophic forgetting, which are common problems when dealing with small amounts of training data. We implemented our system by using the transformers\footnote{\url{https://github.com/huggingface/transformers}} and the adapters\footnote{\url{https://github.com/adapter-hub/adapters}} libraries.

We expected the model to benefit from both mono-lingual and cross-lingual knowledge transfer; therefore, the training process comprised several stages. First, we fine-tuned a Latin language adapter on a publicly available Latin Corpus\footnote{\url{https://github.com/mathisve/LatinTextDataset}} collected from the Latin Library\footnote{\url{https://www.thelatinlibrary.com/}}.
In the next phase of training, we trained a task-specific classification adapter on the English IMDB movie reviews dataset\footnote{\url{https://huggingface.co/datasets/imdb}}. The dataset contains only two labels: positive and negative. We created an adapter with a classification head with four classes, two of which remained unused during this stage.  
Finally, we stacked the task adapter previously trained on English on top of the language adapter, and continued training the task adapter on the annotated data in Latin.

The language adapter was trained for ten epochs with a learning rate 1e-4. For further usage, we took the last checkpoint. The task adapter was trained on data in English for five epochs with a learning rate of 5e-4, and we also took the last checkpoint. Finally, for the submissions, we trained a model on both sets of annotated data for 50 epochs with a 5e-4 learning rate. We used the provided gold annotation example as the validation set for training and measured the F-score on it after each epoch. For submission, we selected the best checkpoint based on the validation F-score.

\section{Results}

\begin{figure*}[h!]
  \centering
  \begin{subfigure}[t]{0.47\textwidth}
        \centering
        \includegraphics[width=\textwidth]{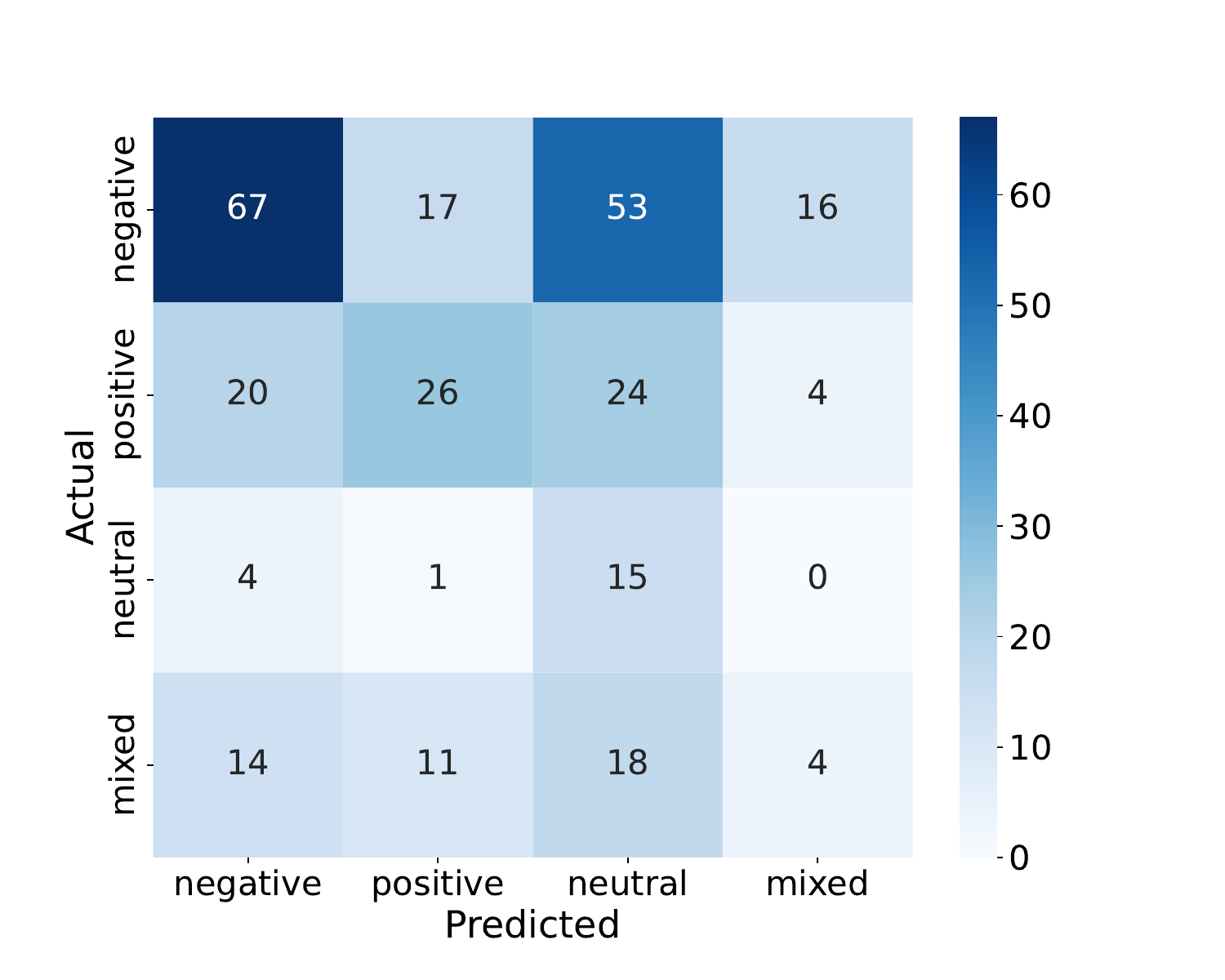}
        \caption{TartuNLP\_1 with lexicon-based heuristic labels.}
        \label{fig:confusion_model1}
    \end{subfigure}%
    ~ 
    \begin{subfigure}[t]{0.47\textwidth}
        \centering
        \includegraphics[width=\textwidth]{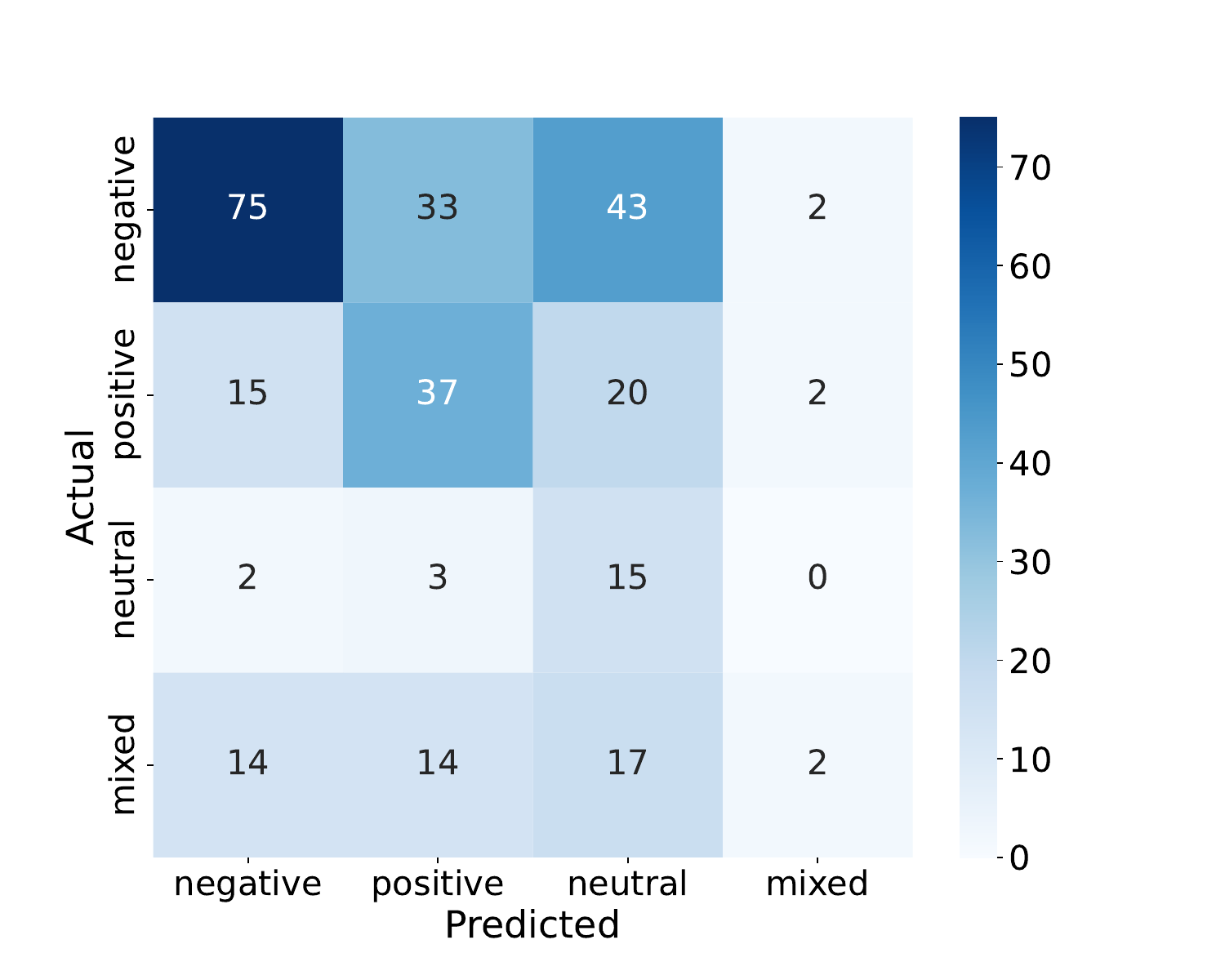}
        \caption{TartuNLP\_2 with GPT4-generated labels.}
        \label{fig:confusion_model1}
    \end{subfigure}
  \caption{Confusion matrices for both submissions.}
\label{fig:confusion}

\end{figure*}

We made two submissions to the Emotion Polarity Detection task; the first one (TartuNLP\_1) fine-tuned on the dataset with the heuristic labels, and the second one (TartuNLP\_2) fine-tuned on the dataset with the LLM-generated labels.
Both submissions obtained competitive results, with the model trained on the LLM-annotated labels (TartuNLP\_2) taking the overall first place and the model trained on the heuristics-annotated data (TartuNLP\_1) taking the second place on micro average F1-score and the third place on the macro average F1-score (see Table~\ref{tab:results}).

\begin{table*}[t]
\centering
\begin{tabular}{lccc}
\toprule
\textbf{Ablation} & \textbf{Micro Avg F1} & \textbf{Macro Avg F1} & \textbf{Val F1} \\
\midrule
Heuristic labels without knowledge transfer        & 0.33 & 0.26 & 0.48 \\
Heuristic labels + Monolingual language transfer     & 0.34 & 0.25 &  0.48 \\
Heuristic labels + Cross-lingual task transfer         & 0.30 & 0.23 & 0.55 \\
Heuristic labels + Both (TartuNLP\_1) & 0.32 & 0.27 &  0.47 \\
\midrule
LLM labels without knowledge transfer     & 0.37 & 0.30 & 0.55 \\
LLM labels + Monolingual language transfer & \textbf{0.38} & \textbf{0.30} & \textbf{0.61} \\
LLM labels + Cross-lingual task transfer & 0.37 & 0.29 & 0.53 \\
LLM labels + Both (TartuNLP\_2) & 0.34 & 0.29 & 0.48 \\
\bottomrule
\end{tabular}
\caption{The results of the ablation study.}
\label{tab:ablation}
\end{table*}

While the scores obtained by the two models are quite close, there is frequent disagreement in their predictions: out of 294 test examples, the models disagreed in 140 examples. In case of disagreement, the heuristics- and LLM-based models made correct predictions in 40 and 57 examples respectively. Meanwhile, in case of agreement, the models correctly predicted the labels of 72 examples out of 154.

The confusion matrices for both models (see Figure~\ref{fig:confusion}) are similar. The models had the most trouble with the mixed class, while the negative class was the easiest to predict; this is in line with findings by~\citet{sprugnoli2023sentiment}, who reported the lowest inter-annotator agreement for the mixed class, while the negative class had the highest agreement, assuming that the test data of the shared task was annotated in a similar manner.

We performed a small ablation study on the labeled test data released by the organizers after evaluating the shared task results to measure the effect of the knowledge transfer methods used:

\begin{enumerate}[label=\arabic*),noitemsep]
    \item Monolingual knowledge transfer from the wider Latin corpus in training the language adapter;
    \item Cross-lingual knowledge transfer from the English IMDB sentiment dataset in training the task adapter.
\end{enumerate}

The results of the study, shown in Table~\ref{tab:ablation}, were somewhat unexpected. 
First of all, we observe that the base model with no knowledge transfer is already as good or better than the submitted models adopting both types of knowledge transfer. Secondly, the monolingual knowledge transfer by training the language adapter improves the micro-averaged F1-score with both types of labels. Finally, the model with the LLM-generated labels benefits more from the monolingual language adapter training resulting in a model that noticeably outperforms our initial submission.

\section{Discussion}

The model with LLM-generated labels obtained better results than the model with lexicon-based heuristic labels, although the final results of both submitted systems are relatively close. However, the ablation study testing the effectiveness of both monolingual and cross-lingual knowledge transfer demonstrated that the model trained on the LLM-annotated data can show even better results when omitting the cross-lingual transfer from English.
This is despite the fact that the number of LLM-annotated examples was nearly twice as small, suggesting that the LLM annotations are of higher quality than the labels based on lexicon-informed heuristics.

Despite our model trained on the LLM-annotated data taking the overall first place, the absolute values are somewhat low and sometimes below the baseline. There might be several reasons related to the choice of the data source and the annotation scheme and procedures.
First, many of the examples appear to be expository or narrative in nature. It is difficult to assign a particular emotive polarity to the texts of that kind.
Furthermore, \citet{sprugnoli2023sentiment} mention that the annotators were instructed to assign labels on the sentence level. However, they were also presented with the wider context of the sentence. This leads us to believe that some labels are actually contextual, especially when the annotated sentence contains only a single word (for example, the sentence "Mentior?" is labeled as mixed).
Secondly, the manual analysis of the examples shows that it is quite difficult to distinguish between mixed and neutral texts. This appears to be true for the trained models, as well.

One possibility of improvement is to reframe the task as a multi-label classification problem instead. The model would be expected to predict the probabilities for the negative and positive labels independently. If the probability of both labels is low, the assigned label can be "neutral"; if both probabilities are high, the label can be "mixed"; otherwise, the label corresponding to the highest probability would be assigned.

\section{Conclusion}
This paper described our solution to the Emotion Polarity Detection task of the EvaLatin Evaluation Campaign. Our submission obtained with a model trained on a dataset with LLM-generated labels achieved the overall first place, showing that LLM-based annotations can be useful for processing texts in Latin. 

\section{Acknowledgments}

This research was supported by the Estonian Research Council Grant PSG721.
\vspace{2cm}

\nocite{*}
\section{Bibliographical References}\label{sec:reference}

\bibliographystyle{lrec-coling2024-natbib}
\bibliography{lrec-coling2024-example}


\end{document}